\documentclass[letterpaper]{article} 
\usepackage{aaai2026}  
\usepackage{times}  
\usepackage{helvet}  
\usepackage{courier}  
\usepackage[hyphens]{url}  
\usepackage{graphicx} 
\usepackage{natbib}  
\usepackage{caption} 
\usepackage{algorithm}
\usepackage{algorithmic}

\usepackage{newfloat}
\usepackage{listings}

\usepackage{url}            
\usepackage{booktabs}       
\usepackage{amsfonts}       
\usepackage{nicefrac}       
\usepackage{microtype}      
\usepackage{xcolor}         
\usepackage{graphicx}
\usepackage{pifont} 
\usepackage{amssymb}
\usepackage{amsmath}
\usepackage{subcaption}
\usepackage{makecell}
\usepackage{multirow}
\usepackage{colortbl} 
\usepackage{tabularx}
\usepackage{enumitem}
\usepackage{diagbox}
\usepackage{xcolor}

\DeclareCaptionStyle{ruled}{labelfont=normalfont,labelsep=colon,strut=off} 
\floatstyle{ruled}
\newfloat{listing}{tb}{lst}{}
\floatname{listing}{Listing}
\title{Hi-EF: Benchmarking Emotion Forecasting in Human-interaction}
\author{
Haoran Wang\textsuperscript{\rm 1}\equalcontrib,
Xinji Mai\textsuperscript{\rm 1}\equalcontrib,
Zeng Tao\textsuperscript{\rm 1}\equalcontrib,
Junxiong Lin\textsuperscript{\rm 1},
Xuan Tong\textsuperscript{\rm 1},
Ivy Pan\textsuperscript{\rm 2},
Shaoqi Yan\textsuperscript{\rm 3},
\\ Yan Wang\textsuperscript{\rm 4}\correspondingauthors,
Shuyong Gao\textsuperscript{\rm 5,6}\correspondingauthors
}
\affiliations{
\textsuperscript{\rm 1}College of Intelligent Robotics and Advanced Manufacturing, Fudan University\\
\textsuperscript{\rm 2}The University of Hong Kong \\
\textsuperscript{\rm 3}School of Electrical and Electronic Engineering, Shanghai Institute of Techonolgy\\
\textsuperscript{\rm 4}East China Normal University\\
\textsuperscript{\rm 5}Carnegie Mellon University\\
\textsuperscript{\rm 6}Keenon Robotics Co., Ltd\\
{\tt\small \texttt{\{hrwang23, xjmai23\}@m.fudan.edu.cn} ztao19@fudan.edu.cn}
}

\usepackage{bibentry}

\begin{document}

\maketitle

\begin{abstract}
Affective Forecasting is an psychology task that involves predicting an individual's future emotional responses, often hampered by reliance on external factors leading to inaccuracies, and typically remains at a qualitative analysis stage. To address these challenges, we narrows the scope of Affective Forecasting by introducing the concept of Human-interaction-based Emotion Forecasting (EF). This task is set within the context of a two-party interaction, positing that an individual's emotions are significantly influenced by their interaction partner's emotional expressions and informational cues. This dynamic provides a structured perspective for exploring the patterns of emotional change, thereby enhancing the feasibility of emotion forecasting. 

\end{abstract}

\section{Introduction}
\label{sec:intro}

Affective computing is a branch of computer science and artificial intelligence that aims to enable computers to recognize, interpret, process, and simulate human emotions. Proposed by Rosalind Picard \cite{picard2000affective}, this field combines psychology, computer science, and cognitive science to imbue machines with emotional intelligence. Current research focuses primarily on calculating and analyzing an individual’s present emotional state using facial expressions, voice, and physiological signals, which is Emotion Recognition (ER)~\cite{lo2020mer,tran2015learning,liu2020saanet,ma2019emotion,li2023cliper,zhang2022tailor,liu2022mafw}. However, despite significant progress in understanding and responding to current emotions, affective computing often overlooks the prediction of future emotions. Affective Forecasting \cite{wilson2003affective} refers to the process by which individuals predict their future emotional reactions to various events. This cognitive mechanism plays a crucial role in decision-making, influencing choices by anticipating potential emotional outcomes. However, empirical studies~\cite{wilson2003affective, gilbert2007prospection, wilson2005affective} have shown that individuals often exhibit significant inaccuracies in their affective forecasts. For instance, Gilbert and Wilson \cite{gilbert2007prospection} highlight the prevalence of “impact bias”, where people tend to overestimate the intensity and duration of their emotional responses to future events. This bias is evident in both positive and negative contexts, such as overestimating happiness from future achievements or distress from anticipated setbacks. Moreover, factors such as focalism, where individuals focus too narrowly on a single event, and underestimating emotional adaptation, contribute to these forecasting errors \cite{wilson2005affective}. Understanding these predictive inaccuracies is critical, as they can lead to suboptimal decision-making in personal, professional, and social contexts. Recent advancements in this field aim to mitigate these biases by exploring the underlying cognitive and emotional mechanisms that drive affective forecasting \cite{wilson2003affective}.

\begin{figure}[t]\centering
	\includegraphics[width=0.5\textwidth]{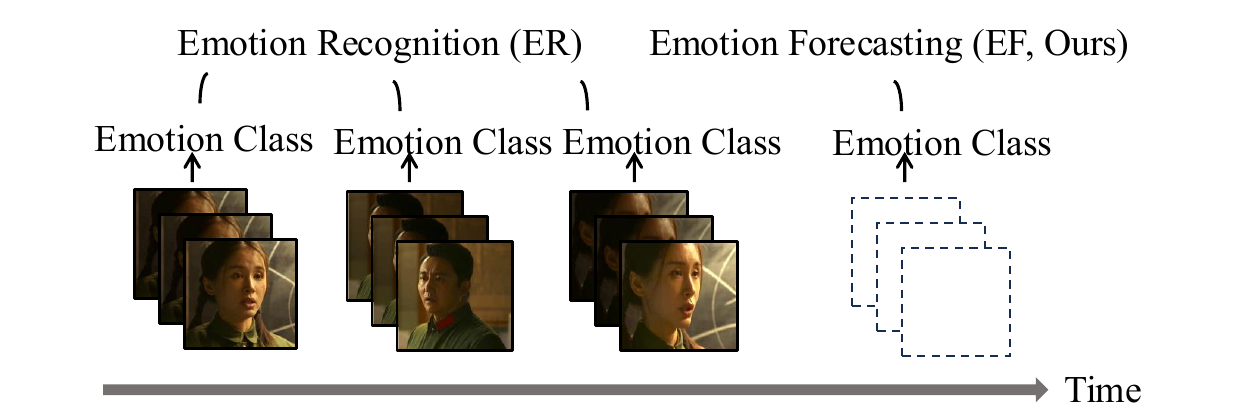}
	\caption{Comparing EF task with ER task.}
    \label{fig10}
\end{figure}

Research on affective forecasting primarily focuses on psychology and economics. However, we aim to redefine the paradigm of affective forecasting to transform this problem into a deep learning challenge. We propose a new task: Emotion Forecasting (EF), predicting one party’s future emotion during a two-party interaction by using short-term contextual information and the current emotional state of the other party. First, emotional exchange is bidirectional: one person’s emotions can affect the other’s, allowing for more accurate predictions of emotional changes \cite{hatfield1993emotional}. Second, two-party interactions occur over short time frames, making the emotional predictions more reliable as responses to recent events are more direct and intense \cite{loewenstein1992anomalies}. Lastly, the dynamic context of interactions provides rich background information, such as conversation content, tone, and emotional shifts, enhancing the precision of affective forecasting \cite{barsade2002ripple}. By remodelling affective forecasting to focus on interpersonal emotional dynamics, we can better understand and predict emotional changes in relationships, providing new tools and methods for emotion management and support \cite{wilson2005affective, gilbert2007prospection}. And we illustrate the difference between ER task and EF task in Figure \ref{fig10}.

Based on these, we list these criteria to construct datasets for solving the EF task:
(1) Multi-modal information: By analyzing facial expressions, conversational content, and vocal intonations, we can better assess a person's emotions, thereby enabling more accurate predictions of the emotions the other party may experience. (2) Strong-Interactive Scenes: Videos should be sourced from highly interactive scenarios, enabling enough amount of interaction data for the dataset. (3) Multilayered-Contextual Information: The data format needs to include three layers of contextual information: contextual information, one party's current emotional state, and the emotion of the other party that needs to be forecasted.

In adherence to these guidelines, we have developed the Human-interaction-based Emotion Forecasting (Hi-EF) dataset. Hi-EF boasts 3,069 Multilayered-Contextual Interaction Samples (MCIS), featuring in 5,242 video clips with abundant affect-related labels. To meet the stringent criteria for data collection, we have devised a comprehensive two-stage strategy to acquire a sufficient pool of candidate MCIS. Furthermore, we have implemented a reliable three-round annotation process to ensure high-quality labeling, given the complexity of the emotional information encapsulated in the labels to make sure the emotion is annotated with high quality. For the challenging EF task, we proposed the baseline method and made comprehensive experiments to demonstrate the feasibility of EF task. Besides that, more experiments of other affective computing tasks are also provided.

Our work makes the following significant contributions:
\begin{itemize}[leftmargin=*]

\item We have transformed Affective Forecasting from a psychological and economic problem into a deep learning task and developed an Emotion Forecasting paradigm based on two-party interactions. This introduces a novel task in the field of affective computing: Emotion Forecasting (EF). Previously, affective computing primarily focused on recognizing current emotions. Our work provides a new perspective by aiming to predict future emotions.
\item To address this novel EF task, we have constructed a Human-interaction-based Emotion Forecasting dataset (Hi-EF). This dataset introduces a unique data format, Multilayered-Contextual Interaction Samples (MCIS), specifically designed for the EF task. MCIS encompasses short-term contextual information, one party's current emotional state, and the other party's future emotional state. It includes three modalities and rich affect-related labels for both parties. Additionally, our dataset can also be utilized for other affective computing tasks.
\item To generate MCIS, we developed a construction procedure including candidate MCIS generation, anomaly detection and multifaceted-assisted reliable annotation.
\item We have designed a paradigm for the EF task and, based on this paradigm, established a baseline for the EF task. Extensive experiments and analyses were conducted on the Hi-EF dataset, demonstrating the feasibility of the EF task and the significance of the Hi-EF dataset. We will soon release our method's code and the dataset for further research and validation.
\end{itemize}

\begin{table*}[htbp]
\centering
\scriptsize
\resizebox{\textwidth}{!}{%
\begin{tabular}{ccccccccccccccc}
\toprule
 &
   &
   &
   &
  \multicolumn{2}{c}{Text Description} &
  &
  \multicolumn{4}{c}{Affective Component} &
   &
   &
   &
   
   \\
   \cmidrule(lr){5-6} \cmidrule(lr){8-11}
     \multirow{-2}{*}{ID} &
\multirow{-2}{*}{Datasets} &
  \multirow{-2}{*}{Class} &
  \multirow{-2}{*}{Scene} &
  Fac. &
  Scene &
  \multirow{-2}{*}{Modality} &
  Polar. &
  Inten. &
  Emo. &
  Uncer. &
  \multirow{-2}{*}{DFER} &
  \multirow{-2}{*}{MER} &
  \multirow{-2}{*}{ERC} &
  \multirow{-2}{*}{\textbf{EF}} 
 \\
  \toprule
  1 &
  IEMOCAP (2008)~\cite{busso2008iemocap} &
  5 &
  1 &
  - &
  - &
  V. \&A.   \&T. &
  - &
  - &
  $\large\checkmark$ &
  - &
  - &
  $\large\checkmark$ &
  - &
  - \\
  2 &
  HOW (2011)~\cite{morency2011towards}  &
  3 &
  1 &
  - &
  - &
  V. \&A.   \&T. &
  - &
  - &
  $\large\checkmark$ &
  - &
  - &
  $\large\checkmark$ &
  - &
  - \\
  3 &
  CMU-MOSEI (2018)~\cite{zadeh2018multimodal} &
  6 &
  1 &
  - &
  - &
  V. \&A.   \&T. &
  - &
  - &
  $\large\checkmark$ &
  - &
  - &
  $\large\checkmark$ &
  - &
  - \\
  4 &
  MELD (2018)~\cite{poria2018meld} &
  7 &
  1 &
  - &
  - &
  V. \&A.   \&T. &
  - &
  - &
  $\large\checkmark$ &
  - &
  - &
  $\large\checkmark$ &
  $\large\checkmark$ &
  - \\
  5 &
  CAER (2019)~\cite{Lee_Kim_Kim_Park_Sohn_2019} &
  7 &
  1 &
  - &
  - &
  V. &
  - &
  - &
  $\large\checkmark$ &
  - &
  $\large\checkmark$ &
  - &
  - &
  - \\
  6 &
  DFEW (2020)~\cite{jiang2020dfew} &
  7 &
  1 &
  - &
  - &
  V. &
  - &
  - &
  $\large\checkmark$ &
  - &
  $\large\checkmark$ &
  - &
  - &
  - \\
  7 &
  FERV39K (2022)~\cite{wang2022ferv39k} &
  7 &
  22 &
  - &
  - &
  V. &
  - &
  - &
  $\large\checkmark$ &
  - &
  $\large\checkmark$ &
  - &
  - &
  - \\
  8 &
  MAFW (2022)~\cite{liu2022mafw} &
  11 &
  1 &
  $\large\checkmark$ &
  $\large\checkmark$ &
  V. \&A.   \&T. &
  - &
  - &
  $\large\checkmark$ &
  - &
  $\large\checkmark$ &
  $\large\checkmark$ &
  - &
  - \\ 
  \rowcolor{pink}
  9 &

  Hi-EF (2025) &
  7 &
  4 &
  $\large\checkmark$ &
  $\large\checkmark$ &
  V. \&A.   \&T. &
  $\large\checkmark$ &
  $\large\checkmark$ &
  $\large\checkmark$ &
  $\large\checkmark$ &
  $\large\checkmark$ &
  $\large\checkmark$ &
  $\large\checkmark$ &
  $\large\checkmark$ \\
  \bottomrule
\end{tabular}%
}
\caption{Summary of existing Affective Computing datasets (ID:1-4), ERC dataset (ID:4) and DFER datasets (ID:4-8) and our built Hi-EF. (Fac. = Facial movements; Polar. = Interaction polarity; Inten. = Interaction intensity; Emo. = Emotion; Uncer. = Uncertainty; V. = Video; A. = Audio; T. = Text.}
\label{tab1}
\end{table*}
\section{Significance of EF Task}
Unlike traditional Affective Computing tasks that focus on recognizing current emotions, the EF task aims to predict future potential emotions based on interactional context. Accurate and reasonable predictions in this domain have a wide array of applications. We outline several future research directions and applications:
\begin{itemize}[leftmargin=*]
\item \textbf{Individual Emotion Modeling.} By analyzing extensive data from a person's interactions, we can model their emotional responses, identifying what types of interactions are likely to elicit specific emotions. This modeling is useful in the medical field for treating conditions such as depression, enabling personalized therapeutic approaches.

\item \textbf{Anthropomorphic Emotion Generation.} In an interactive scenario, by evaluating the emotional state and contextual information of the other party, we can predict the appropriate emotional response a human should exhibit. This ability enables the generation of anthropomorphic emotions, which is advantageous for developing emotionally intelligent robots. Such robots can respond in a more human-like and empathetic manner, thereby enhancing their effectiveness in roles that require human interaction.
\end{itemize}

\section{Relevant Datasets of Emotion Forecasting: Emotion Recognition Datasets}
\label{sec:formatting}
Since the EF task is a novel task and there are no existing EF datasets, we will introduce the ER datasets that are most relevant to the EF task to aid in understanding our Hi-EF dataset. These datasets are presented in Table \ref{tab1}. By presenting these datasets, we aim to facilitate a better understanding of the distinctions between our Emotion Forecasting task, the Hi-EF dataset, and traditional Emotion Recognition tasks and datasets. This comparison will help clarify the unique aspects of our work.
ER datasets have two mainstream types: video-driven ER datasets and multi-modal ER datasets. The primary classification is based on the dominant modalities that constitute the datasets and their corresponding related tasks, including Dynamic Facial Expression Recognition (DFER), Multi-modal Emotion Recognition (MER), and Emotion Recognition in Conversation (ERC).

\textbf{Video-driven Emotion Recognition Datasets.} 
Video-driven Emotion Recognition datasets are dedicated to identifying human emotions from dynamic visual information. These datasets includes DFER datasets, typically originating from controlled laboratory settings, as exemplified by the CK+~\cite{lucey2010extended} and Oulu-CASIA~\cite{zhao2011facial} datasets. Additionally, there are DFER datasets in the wild, such as AFEW~\cite{dhall2018emotiw}, Aff-wild~\cite{kollias2019deep}, AFEW-VA~\cite{kossaifi2017afew}, CAER~\cite{lee2019context}, DFEW~\cite{jiang2020dfew}, FERV39k~\cite{wang2022ferv39k}. Datasets like MAFW~\cite{liu2022mafw}, which offer diverse modalities and emotion labels while is still video-driven, since the annotation is mainly based on the visual modality, while it could also be used in MER task. 
\\ \textbf{Multi-modal Emotion Recognition Datasets.} 
MER datasets focus not just the facial expression. Instead, they pay more attention to fuse the different modalities to analyze the whole emotional state of a person. And the class is not only the expression, but also sentiment including positive, neutral and negative and Activation-Valence-Dominance like IEMOCAP~\cite{busso2008iemocap}, HOW~\cite{morency2011towards}, CMU-MOSEI\cite{zadeh2018multimodal} and MELD~\cite{poria2018meld}. Especially for MELD, it is text-based Emotion Recognition dataset designed for ERC.

The EF task differs from the ER task, although it does involve elements of ER. In the EF task, we first need to recognize one party's emotions and then, using the contextual information, predict the potential emotions of the other party. This task structure results in the Hi-EF dataset's MCIS format being entirely different from typical ER formats. For example, in DFER tasks, the dataset usually consists of single videos labeled with the expressions of the person in the video, typically using a seven-category classification for expressions. In contrast, MCIS includes information from both parties' interactions. It comprises four consecutive videos: the first two videos provide contextual information, the third video presents one party's emotional state, and the fourth video is where we need to predict the other party's emotional state. More details about the EF task, Hi-EF, and MCIS will be elaborated in Section \ref{section3}.

\section{The Hi-EF Dataset}
\label{section3}
\subsection{Paradigm of Emotion Forecasting }

\begin{figure*}[t]
        \centering
       \includegraphics[width=\textwidth]{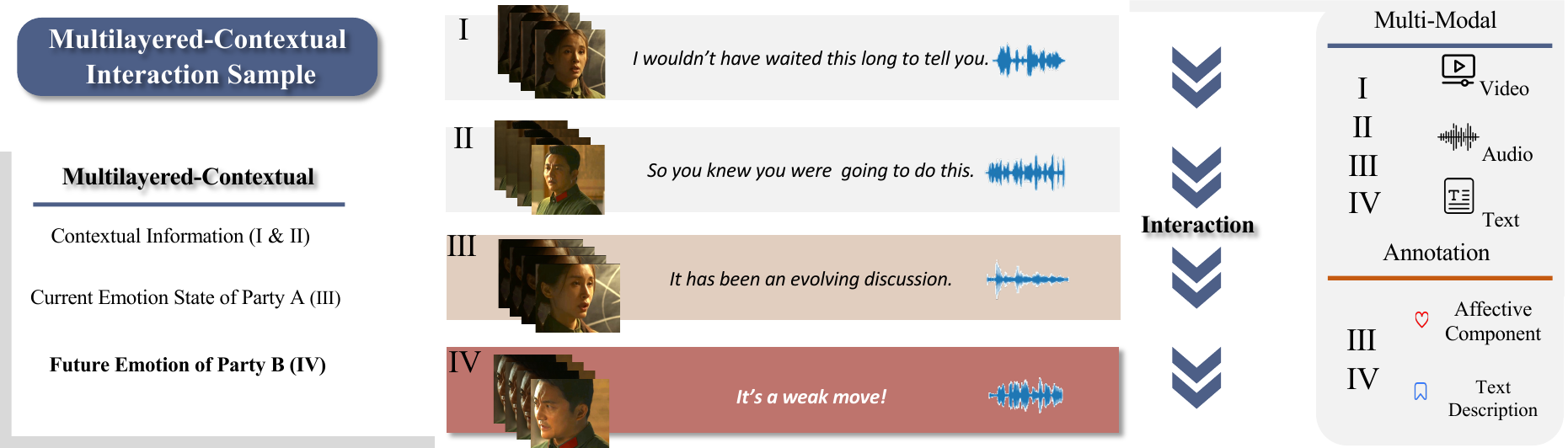}
       \caption{Overview of MCIS: In order to realize the task of forecasting emotion during the interaction process, we design a new form of data MCIS with multilayered-contextual information which includes contextual information (clip \uppercase\expandafter{\romannumeral1} and \uppercase\expandafter{\romannumeral2}), current emotion state of Party A (clip \uppercase\expandafter{\romannumeral3}) and future emotion of Party B (clip \uppercase\expandafter{\romannumeral4}). To better evaluate Party A’s emotional state and more accurately predict Party B’s emotions, we provide three modalities for MCIS: video, audio and text.}

       \label{Fig1}
\end{figure*}
The EF task entails leveraging contextual information from the interaction between two parties and the emotions of one party to predict the potential future emotions of the other. Imagine a scenario where Party A shares a joyful experience with Party B, who shows significant interest. If A displays happiness during this interaction, it is highly probable that B will also exhibit happiness subsequently. This scenario can be divided into three layers. The first layer encompasses contextual information, serving as background and supplementary data. The second layer pertains to Party A's emotional state, which directly influences Party B's emotions. The third layer comprises Party B's own emotions, which are what we aim to forecast. Therefore, we believe the EF task should incorporate the following three functions: understanding contextual information, recognizing Party A's emotional state, and integrating these two layers of information for forecasting. These functions are defined as follows:
\begin{itemize}[leftmargin=*]
    \item Contextual Interaction Fusion Function \(I\) = \(f_{CI}(\cdot)\): This function merges various layers of contextual information, providing a refined backdrop for emotion interpretation and forecasting. \(I\) indicates contextual information, such as non-verbal cues and the level of participation of the participants.

    \item Current Party Emotion Recognition Function \(E_{A} = f_{ER}(\cdot)\): It recognizes Party~A's current emotional state from multi-modal cues.

    \item Future Emotion Forecasting Function \(E_{B}\) = \(f_{EF}(I, E_A)\): Leveraging insights from \(f_{CI}\) and \(f_{ER}\), this function forecasts the probable future emotion of Party B. \(E_B\) is the future emotion of Party B which will be forecasted.

\end{itemize}

It is worth noting that in our model architecture, the three functions are not explicitly defined as independent modules but are integrated into the overall design, which still follows the proposed paradigm.

\subsection{Design of Multilayered-Contextual Interaction Sample (MCIS)}
To adapt to the EF task, we needed to define a data format. In the introduction, we also outlined our requirements for the dataset: Multi-modal, Strong-Interactive Scene, and Multilayered-Contextual information. Therefore, we designed the \textbf{Multilayered-Contextual Interaction Sample (MCIS)} as the data format for the Hi-EF dataset. Figure 2 illustrates MCIS. An MCIS primarily consists of the following components: Contextual Information: Video clip \uppercase\expandafter{\romannumeral1} and \uppercase\expandafter{\romannumeral2}; Current Information of Party A: Video clip \uppercase\expandafter{\romannumeral3}; Forecasted Information of Party B: Video clip \uppercase\expandafter{\romannumeral4}. We provide three modalities for each video clip: video, text (utterance), and audio. Additionally, we provide annotation information for clips \uppercase\expandafter{\romannumeral3} and \uppercase\expandafter{\romannumeral4}.
It is important to note that for the Contextual Information part, we only use two clips. We did not choose more than two videos for contextual information mainly due to the following considerations: 1) In Affective Forecasting~\cite{frederick2002time,loewenstein1992anomalies,ainslie1975specious,laibson1997golden}, overly long contextual information might introduce noise, affecting the affective forecast. 2) Excessively long contextual information significantly increases computational costs. 3) Contextual information serves as supplementary data, while the primary data for EF is the current emotional state of Party A \cite{loewenstein1992anomalies}. 

\begin{figure*}[!t]\centering
	\includegraphics[width=\textwidth]{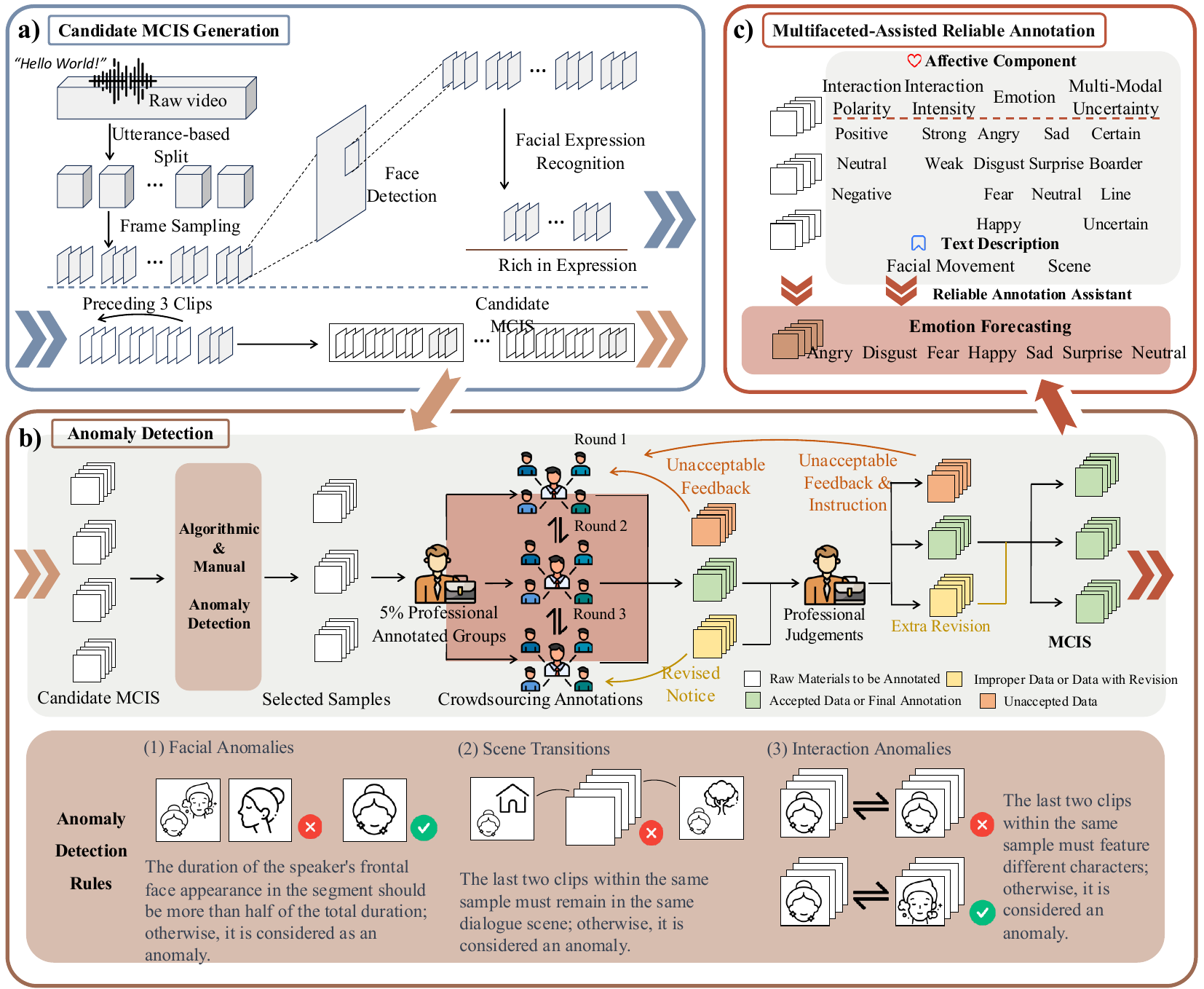}
	\caption{Overview of the Construction Procedure for the Hi-EF Dataset. a) Shows the process of generating candidate MCIS. b) Illustrates the anomaly detection and the rules and multi-facet assisted reliable annotation for candidate MCIS. c) Details the annotation process. All other annotation information is intended to support the more accurate labeling of emotions. Naturally, we will provide all the annotation details within the dataset.}
    \label{fig3}
\end{figure*}

\subsection{Dataset Construction Procedure}
In this section, we present an overview of our database construction process, which consists of three stages: raw data collection and candidate clip generation, candidate MCIS creation, and MCIS annotation.
\\ \textbf{Data Collection and Candidate MCIS Generation.} 
We source raw videos from shows rich in interactive scenes and extract subtitles to segment them into clips based on utterance timestamps. 
Since the initial clips contain noise, we refine them using facial detection and a state-of-the-art ER model~\cite{tao20243} to isolate emotionally relevant segments for MCIS construction. 
Each MCIS is formed by combining three preceding clips with the emotionally significant one, as illustrated in Figure~\ref{fig3}~a). 
This process yields a large number of candidate MCIS samples.
\\ \textbf{Anomaly Detection Rules.} 
Our utterance-based MCIS generation strategy occasionally yields abnormal clips. 
To address this, we design an anomaly detection mechanism to ensure annotation quality. 
As illustrated in Figure~\ref{fig3}~b), we define three anomaly types: 
(1) \textit{Facial anomalies} — clips lacking a frontal face view of the speaker for at least half their duration; 
(2) \textit{Scene transitions} — MCIS whose final two clips belong to different scenes; 
(3) \textit{Interaction anomalies} — MCIS featuring the same individual in the last two clips. 
We implement an algorithm to pre-screen the first and third types, filtering out about half of the invalid MCIS. 
Given the limitations of automatic detection under strict criteria, manual inspection remains indispensable. 
Therefore, we combine algorithmic and manual detection to enhance overall accuracy, which also aligns with the subsequent annotation step. 
\\ \textbf{Multifaceted-Assisted Reliable Annotation.} 
The Hi-EF database adopts a novel data format and undergoes a rigorous three-round manual annotation process (Figure~\ref{fig3}~b–c) involving video, audio, and text modalities. 
In the first round, crowdsourced workers apply anomaly criteria to select valid MCIS. 
In the second round, professional annotators validate and refine these samples, ensuring consistency and annotating affective components and textual descriptions. 
The final round re-assesses the double-checked MCIS, focusing on the last two clips. 
Each MCIS is labeled by five independent annotators, and a professor reviews these annotations with multifaceted auxiliary information to determine or refine the final label. 
We provide comprehensive annotation details, including textual descriptions (facial actions, scene information) and emotional components (interaction polarity, intensity, emotion categories, and uncertainty). 
These annotations collectively support reliable emotion categorization and can facilitate other affective computing tasks. 
Detailed annotation guidelines are provided in the \textbf{Appendix}.

\subsection{Dataset Statistics}
The Hi-EF dataset has been meticulously compiled from TV shows, amassing over 100 hours of content and dividing it into more than 60,000 video clips. Following preprocessing, 30,000 clips containing emotion have been retained, and through a series of data refinement, a half of the candidates are eliminated. Manual judgment has further refined the dataset, resulting in a selection of 3,069 MCIS and 5,242 video clips with high-quality annotations. Additional dataset statistics are provided in the \textbf{Appendix}

\section{Experiment}
\label{section4}

\subsection{Experiment Setup}
\textbf{Hi-EF Protocol.} To establish a robust benchmark for the EF task within the Hi-EF dataset, we partitioned the dataset, comprising 3,069 MCIS, into training (70$\%$) and testing (30$\%$) sets, with the training set further segmented into a validation subset. It is particularly important to note that, unlike traditional data formats, randomly splitting the training and test sets by proportion may result in overlapping clips between MCISs, leading to data leakage. Therefore, we implemented additional criteria to avoid this situation. For the EF task, our experiments unfolded as follows: We employed 2,150 MCIS for training. Subsequently, we use different encoders and fusion strategies for experiments. 
\\ \textbf{Implementation Details.}
All experiments were implemented using PyTorch on Nvidia RTX 3090 GPUs. We set the learning rates to be in the range of 1e-2 to 1e-5, weight decay to 1e-6, and used a batch size of 8. All models were trained from scratch using optimizer (Adam) for 50 epochs, with uniform frame subsampling at an interval of 8 frames.
Furthermore, due to the limited availability of data in the context of Hi-EF, we employed several data augmentation techniques, including random cropping, changes in lighting conditions, and image flipping. To reduce the computational demands, all cropped images were resized to 112$\times$112, and the entire images were resized to 112$\times$168.
\\ \textbf{Evaluation Metrics.} Drawing inspiration from prevalent Affective Computing tasks~\cite{schuller2010cross,jiang2020dfew,liu2022mafw,wang2022ferv39k,wang2022dpcnet,lee2019context,zhang2017facial}, we selected two traditional metrics for evaluation: Weighted Average Recall (WAR) and Unweighted Average Recall (UAR).

\begin{figure}[!t]
\centering
	\includegraphics[width=0.47\textwidth]{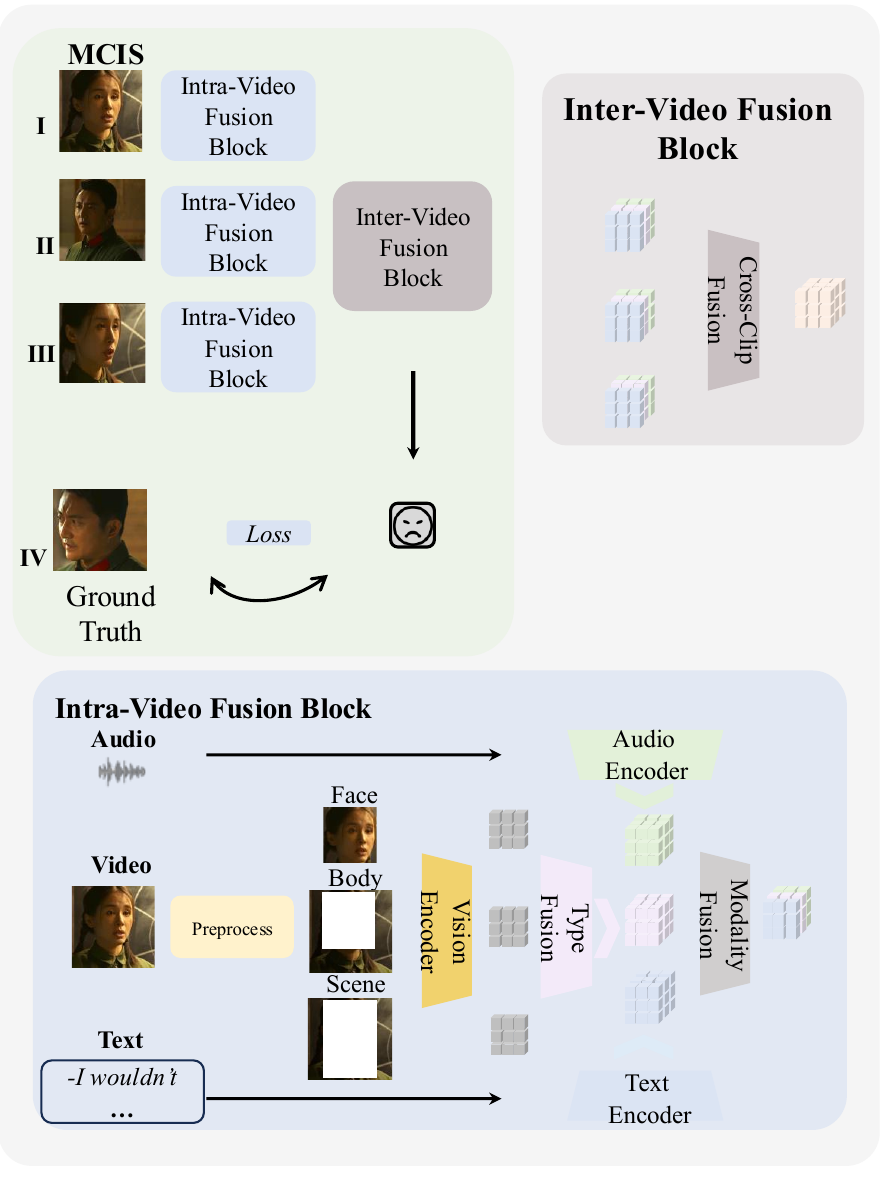}
	\caption{Overview of our model architecture. The first three clips in MCIS are respectively input into the intra-video fusion block, where the obtained features are then fed into the inter-video fusion block for prediction. The final prediction result is compared with the emotion label of clip \uppercase\expandafter{\romannumeral4} in MCIS to calculate the loss. In the intra-video fusion block, video frames are divided into facial, posture, and scene parts, which are separately input into the vision encoder. These features are then fused to form the video modality information. Meanwhile, audio and text are input into the audio encoder and text encoder, respectively. After obtaining the features from the three modalities, they are fused to produce the feature representation of a single video. The inter-video fusion block primarily fuses the features of the three clips.}
    \label{fig6}
\end{figure}
\noindent \textbf{Model Architecture.} Our model architecture, as depicted in Figure \ref{fig6}, is divided into two main parts: intra-video fusion block and inter-video fusion block. Firstly, intra-video fusion block is further divided into two segments. One clip involves the alignment and fusion of the three modalities: video, audio, and text. The other clips focuses on the video modality, where we divide video frames into three parts: facial expressions, body posture, and scene information, as all three contribute to the overall emotional state of a person. For the vision encoder, we use ResNet18\cite{he2016deep}, I3D\cite{carreira2019short}, and ViT\cite{dosovitskiy2020image}. For the audio encoder, we employ the commonly used AudioCLIP\cite{guzhov2022audioclip} Encoder, and for the text encoder, we use the CLIP\cite{radford2021learning} text encoder. Regarding the intra-video fusion strategy, we adopt two approaches: one is the average method, and the other is fusion using a transformer. For inter-video fusion block, we primarily use average, transformer, and a combination of LSTM\cite{graves2012long} and Transformer\cite{vaswani2017attention}.

\begin{table*}[t]
\centering

\resizebox{\textwidth}{!}{
\begin{tabular}{cccccccccc}
\toprule
\multicolumn{4}{c}{Video} & \multirow{2}{*}{Text} & \multirow{2}{*}{Audio} & \multirow{2}{*}{Inter Fusion} & \multirow{2}{*}{WAR} & \multirow{2}{*}{UAR} \\ \cline{1-4} \specialrule{0em}{1pt}{2pt}
Face & Body & Scene & Intra Fusion & & & & & \\ \hline \specialrule{0em}{1pt}{2pt}
\multicolumn{9}{l}{\textit{\textbf{Resnet18 Backbone}}} \\
R18~\cite{he2016deep} & \ding{55} & \ding{55} & \ding{55} & \ding{55} & \ding{55} & Average & 22.22 & 17.02 \\
R18 & R18 & R18 & Average & \ding{55} & \ding{55} & Average & 23.15 & 18.58 \\
R18 & R18 & R18 & Transformer & \ding{55} & \ding{55} & Average & 32.41 & 19.75 \\ 
R18 & R18 & R18 & Average & \ding{55} & \ding{55} & Transformer & 28.70 & 21.09 \\ \hline \specialrule{0em}{1pt}{2pt}
\multicolumn{9}{l}{\textit{\textbf{I3D Backbone}}} \\
I3D\cite{carreira2019short} & \ding{55} & \ding{55} & \ding{55} & \ding{55} & \ding{55} & Average & 28.57 & 21.73 \\
I3D & I3D & I3D & Average & \ding{55} & \ding{55} & Average & 29.62 & 21.16 \\
I3D & I3D & I3D & Transformer & \ding{55} & \ding{55} & Average & 29.63 & 21.19 \\ 
I3D & I3D & I3D & Average & \ding{55} & \ding{55} & Transformer & 29.62 & 22.99 \\ \hline \specialrule{0em}{1pt}{2pt}
\multicolumn{9}{l}{\textit{\textbf{ViT Backbone}}} \\
ViT~\cite{dosovitskiy2020image} & ViT & ViT & Transformer & \ding{55} & \ding{55} & LSTM+Transformer & 31.37 & 22.39 \\
ViT & ViT & ViT & Transformer & \ding{55} & AudioCLIP & LSTM+Transformer & 33.34 & 21.27 \\ 
ViT & ViT & ViT & Transformer & CLIP & \ding{55} & LSTM+Transformer & 31.03 & 21.80 \\ 
ViT & ViT & ViT & Average & CLIP & AudioCLIP & Average & \multicolumn{1}{l}{31.44} & \multicolumn{1}{l}{18.55} \\
ViT & ViT & ViT & Transformer & CLIP & AudioCLIP & Average &  33.38& 21.71 \\ 
ViT & ViT & ViT & Average & CLIP & AudioCLIP & LSTM+Transformer & \underline{34.49} & \underline{23.19} \\
 \rowcolor{lightgray}ViT & ViT & ViT & Transformer & CLIP & AudioCLIP & LSTM+Transformer & \textbf{35.19} & \textbf{23.72}
\\ \bottomrule
\end{tabular}
}

\caption{The comparison of EF experiments result($\%$) with different modalities, backbones, and intra and inter fusion strategies. \textbf{Bold} indicates the best performance, while \underline{underlining} indicates the second-best performance.}
\label{tab3}
\end{table*}
\begin{table*}[!t]
\centering

\begin{tabular}{cccccccccc}
\toprule
Input & Video & Intra Fusion & Text & Audio &Inter Fusion & WAR & UAR \\ \midrule
\uppercase\expandafter{\romannumeral1}    &\multirow{3}{*}{ViT}&\multirow{3}{*}{Transformer}&\multirow{3}{*}{CLIP}&\multirow{3}{*}{AudioCLIP}& \multirow{3}{*}{\textbackslash} & 29.50   & 19.38   \\ 
\uppercase\expandafter{\romannumeral2}    &      &&&& & 30.89   & 20.55   \\
\uppercase\expandafter{\romannumeral3}    &      &&&& & 31.44   & 20.80   \\ \toprule
\uppercase\expandafter{\romannumeral1} + \uppercase\expandafter{\romannumeral2}  &\multirow{3}{*}{ViT}&\multirow{3}{*}{Transformer}&\multirow{3}{*}{CLIP}&\multirow{3}{*}{AudioCLIP}& \multirow{3}{*}{LSTM + Transformer}           & 31.86   & 20.93   \\
\uppercase\expandafter{\romannumeral1} + \uppercase\expandafter{\romannumeral3}   &    &&&&   & 32.41   & 21.13   \\
\uppercase\expandafter{\romannumeral2} + \uppercase\expandafter{\romannumeral3}    &   &&&&   & \underline{32.55}   & \underline{21.39}   \\ \toprule
 \rowcolor{lightgray}\uppercase\expandafter{\romannumeral1} + \uppercase\expandafter{\romannumeral2} + \uppercase\expandafter{\romannumeral3}  & ViT & Transformer & CLIP & AudioCLIP & LSTM + Transformer & \textbf{35.19} & \textbf{23.72}   \\ \bottomrule
\end{tabular}%

\caption{Different input strategies of EF experiments results($\%$) on Hi-EF. \textbf{Bold} indicates the best performance, while \underline{underlining} indicates the second-best performance.}
\label{tab7}

\end{table*}
\subsection{Experimental Results}
\textbf{Experimental Results of EF Task.}
We present the experimental results of the EF task in Table~\ref{tab3}. Our main goal is to study the impact of different vision encoders and intra-/inter-video fusion strategies on EF performance. We first compare the vision encoders. Since ResNet18 and I3D cannot be aligned with the other two modalities, we only report video-only results for them, using simple fusion strategies; consequently, they do not achieve competitive performance. We then focus on ViT-based models and analyze the effect of modalities. The results show that the video modality contributes most to emotion analysis. Adding text alone degrades performance, as the text is treated as noise, whereas adding audio enables the model to better interpret the textual information. Using all three modalities improves WAR by 3.82\% and UAR by 1.33\% over the video-only setting, indicating that our modality choice is appropriate for the EF task.
\\ \textbf{Effect of Intra-Video Fusion.}
During two-party interactions, emotional influence arises not only from facial expressions but also from body posture and the surrounding scene. We therefore compare using only facial information with using the combination of face, body, and scene. The combined setting yields higher prediction accuracy, confirming that these factors jointly affect emotional changes. We also study different fusion strategies for these three types of information. The results show that transformer-based fusion clearly outperforms simple averaging, as it can focus on the most informative emotional cues.
\\ \textbf{Effect of Inter-Video Fusion.}
Since MCIS contains multiple layers of information, an appropriate inter-video fusion strategy is crucial. We compare three strategies: average, Transformer, and LSTM+Transformer. LSTM+Transformer performs best, where LSTM captures temporal dependencies among the three clips and the transformer highlights key signals that drive emotional changes. Their combination leads to the most accurate prediction of the other party's potential emotions.
\\ \textbf{Effect of MCIS.}
We conduct ablation experiments on MCIS to study the impact of different input configurations, as reported in Table~\ref{tab7}. With a single input clip, clip~\uppercase\expandafter{\romannumeral3} achieves the best results. With two clips, the combination of clips~\uppercase\expandafter{\romannumeral2} and~\uppercase\expandafter{\romannumeral3} performs best, indicating that more recent information is more beneficial for EF. When three clips are used, the performance surpasses that of the \uppercase\expandafter{\romannumeral2}+\uppercase\expandafter{\romannumeral3} setting, with WAR and UAR improvements of 2.64\% and 2.33\%, respectively, demonstrating the effectiveness of the MCIS design.
\\ \textbf{Derived Tasks of Hi-EF.}
Beyond EF, the Hi-EF dataset can also support other affective computing tasks, such as DFER, MER, and ERC. The corresponding experimental results are reported in the \textbf{Appendix}.

\section{Conclusions and Discussion}

\label{section 6}
In this paper, we have transformed the task of Affective Forecasting from a psychological and economic problem into a deep learning challenge. We define a new task called Emotion Forecasting, which focuses on forecasting emotions during two-party interactions. To facilitate this task, we propose the Hi-EF dataset and a baseline. The dataset features a novel data format, MCIS, which consists of multilayered-contextual interactive information between two parties. We provide rich emotion-related annotations and three modalities for each MCIS. Our primary contribution lies in designing the EF task and creating a suitable dataset and data format to accommodate this task, ultimately demonstrating its feasibility. We believe that the EF task and the Hi-EF dataset will provide a new perspective for future research in affective computing.
\\ \textbf{Limitations.} Our baseline model does not employ a wide variety of model architectures, and the fusion strategies used are simple and commonly adopted methods without much innovation. Our primary contribution lies in designing the EF task and creating a suitable dataset and data format to accommodate this task, ultimately demonstrating its feasibility. We believe that the EF task and the Hi-EF dataset will provide a new perspective for future research in affective computing. In the future, we plan to expand the Hi-EF dataset, test more models, and continuously provide experimental results.

\section{Acknowledgements}
This work was supported by National Natural Science Foundation of China (No.62576109, 62072112), National Key Research and Development Program of China (2023YFC3604802) and National Natural Science Foundation of China (No. 62406075)
\clearpage

\bibliography{main}
\clearpage


\makeatletter
\@ifundefined{isChecklistMainFile}{
  \newif\ifreproStandalone
  \reproStandalonetrue
}{
  \newif\ifreproStandalone
  \reproStandalonefalse
}
\makeatother

\ifreproStandalone
\documentclass[letterpaper]{article}
\usepackage[submission]{aaai2026}
\setlength{\pdfpagewidth}{8.5in}
\setlength{\pdfpageheight}{11in}
\usepackage{times}
\usepackage{helvet}
\usepackage{courier}
\usepackage{xcolor}
\frenchspacing

\begin{document}
\fi
\setlength{\leftmargini}{20pt}
\makeatletter\def\@listi{\leftmargin\leftmargini \topsep .5em \parsep .5em \itemsep .5em}
\def\@listii{\leftmargin\leftmarginii \labelwidth\leftmarginii \advance\labelwidth-\labelsep \topsep .4em \parsep .4em \itemsep .4em}
\def\@listiii{\leftmargin\leftmarginiii \labelwidth\leftmarginiii \advance\labelwidth-\labelsep \topsep .4em \parsep .4em \itemsep .4em}\makeatother

\setcounter{secnumdepth}{0}
\renewcommand\thesubsection{\arabic{subsection}}
\renewcommand\labelenumi{\thesubsection.\arabic{enumi}}

\newcounter{checksubsection}
\newcounter{checkitem}[checksubsection]

\newcommand{\checksubsection}[1]{%
  \refstepcounter{checksubsection}%
  \paragraph{\arabic{checksubsection}. #1}%
  \setcounter{checkitem}{0}%
}

\newcommand{\checkitem}{%
  \refstepcounter{checkitem}%
  \item[\arabic{checksubsection}.\arabic{checkitem}.]%
}
\newcommand{\question}[2]{\normalcolor\checkitem #1 #2 \color{blue}}
\newcommand{\ifyespoints}[1]{\makebox[0pt][l]{\hspace{-15pt}\normalcolor #1}}

\ifreproStandalone
\end{document}
\fi

\end{document}